\pdfoutput=1

\documentclass[11pt]{article}

\usepackage{naacl2021}

\usepackage{times}
\usepackage{latexsym}

\usepackage[T1]{fontenc}

\usepackage[utf8]{inputenc}

\usepackage{microtype}
\usepackage{graphicx}
\usepackage{eurosym}
\usepackage{tablefootnote}
\graphicspath{ {images/} }
\usepackage{hyperref}
\usepackage{textcomp}

\usepackage{hhline}
\usepackage[utf8]{inputenc}

\usepackage[T1]{fontenc}
\usepackage{listings}
\usepackage{stfloats}
\usepackage{csquotes}
\usepackage{array}
\usepackage{booktabs}

\usepackage[super]{nth}
\usepackage[inline,shortlabels]{enumitem}
\usepackage{multirow}
\usepackage{makecell}
\usepackage{siunitx}
\usepackage{multicol}
\usepackage{graphicx}

\newcounter{notecounter}
\newcommand{\enotesoff}{\long\gdef\enote##1##2{}}
\newcommand{\enoteson}{\long\gdef\enote##1##2{{
			\stepcounter{notecounter}
			\large\bf
			\hspace{100cm}\arabic{notecounter} $<<<$ ##1: ##2
			$>>>$\hspace{1cm}}}}
\enoteson
\enotesoff

\usepackage{svg}
\usepackage{float}
\usepackage{ amssymb }
\usepackage{amsmath,amsfonts,amssymb}

\title{Improving the Lexical Ability of Pretrained Language Models \\ for Unsupervised Neural Machine Translation
}

\author{ 
	Alexandra Chronopoulou,
    Dario Stojanovski,
	Alexander Fraser\\\\
	Center for Information and Language Processing, LMU Munich, Germany \\
	{\tt \{achron, stojanovski, fraser\}@cis.lmu.de}
	}

\begin{document}
\maketitle
\begin{abstract}
Successful methods for unsupervised neural machine translation (\textsc{unmt}) employ cross-lingual pretraining via self-supervision, often in the form of a masked language modeling or a sequence generation task, which 
 requires the model to align the lexical- and high-level representations of the two languages. 
While cross-lingual pretraining works for similar languages with abundant corpora, it performs poorly in low-resource and distant languages.
Previous research has shown that this is because the representations are not sufficiently aligned.
In this paper,
we enhance the
bilingual masked language model pretraining with lexical-level information
by
using type-level 
cross-lingual subword embeddings.
Empirical results demonstrate improved performance both on \textsc{unmt} (up to $4.5$ BLEU) and bilingual lexicon induction using our method compared to a
\textsc{unmt} baseline. 


\end{abstract}

\section{Introduction}

\textsc{\textsc{unmt}} is an effective approach for translation without parallel data.
Early approaches 
transfer information from static pretrained cross-lingual embeddings 
 to the encoder-decoder model to provide an implicit bilingual signal \cite{lample2017unsupervised,artetxe2017unsupervised}.
\citet{lample2019cross} suggest to instead pretrain
a bilingual language model (\textsc{xlm}) and use it to initialize \textsc{unmt}, as it can
successfully encode higher-level text representations. This approach largely improves translation scores for language pairs with plentiful monolingual data.
However, while \textsc{\textsc{unmt}} is effective for high-resource languages, 
it yields poor results when one of the two languages is low-resource \cite{guzman2019flores}.
\citet{marchisio2020does} show that there is a strong correlation between bilingual lexicon induction (BLI) and final translation performance when using pretrained cross-lingual embeddings, converted to  phrase-tables, as initialization of a \textsc{unmt} model \cite{artetxe-etal-2019-effective}. 
\citet{Vulic2020ProbingPL} observe that static cross-lingual embeddings achieve higher BLI  scores compared to multilingual language models (\textsc{lm}s), meaning that they obtain a better lexical-level alignment. Since bilingual \textsc{lm} pretraining is an effective form of 
initializing a \textsc{unmt} model, improving the overall representation of the masked language model (\textsc{mlm}) is essential to
obtaining
a higher translation performance.

In this paper, we propose a new method to enhance the embedding alignment of a bilingual language model, entitled \textit{lexically aligned \textsc{mlm}}, that serves as initialization for \textsc{unmt}. Specifically, we learn type-level embeddings separately for the two languages of interest.
We map
these monolingual embeddings to a common space and use them to initialize the embedding layer of an \textsc{mlm}. Then, we train the \textsc{mlm} on both languages.
Finally, we
transfer the trained model to the encoder and decoder of an \textsc{nmt} system. We train the \textsc{nmt} system in an unsupervised way. We outperform a \textsc{unmt} baseline and demonstrate the importance of cross-lingual mapping of token-level representations. We also conduct an analysis to investigate the correlation between BLI,  1-gram precision and translation results. We finally investigate whether cross-lingual embeddings should be updated or not during the \textsc{mlm} training process, 
in order to preserve lexical-level information useful for \textsc{unmt}.
We make the code used for this paper publicly available\footnote{ \url{https://github.com/alexandra-chron/lexical_xlm_relm}}.

\section{Proposed Approach}
Our approach has three distinct steps, which are described in the following subsections.

\subsection{\textit{VecMap} Embeddings} 
Initially, we split the monolingual data from both languages using BPE tokenization \cite{sennrich2015neural}. We build \textit{subword} monolingual embeddings with \textit{fastText} \cite{bojanowski-etal-2017-enriching}. 
Then, we map the monolingual embeddings of the two languages to a shared space, using \textit{VecMap} \cite{artetxe-etal-2018-robust}, with identical tokens occurring in both languages serving as the initial seed dictionary, as we do not have any bilingual signal. This is different from the original \textit{VecMap} approach, which operates at the \textit{word level}. We use the mapped embeddings of the two languages to initialize the embedding layer of a Transformer-based encoder  \cite{vaswani2017attention}. 

\subsection{Masked Language Model Training} \label{sec:mlm}
\begin{figure*}[t]
	\centering
	\includegraphics[width=0.9\textwidth, page=1]{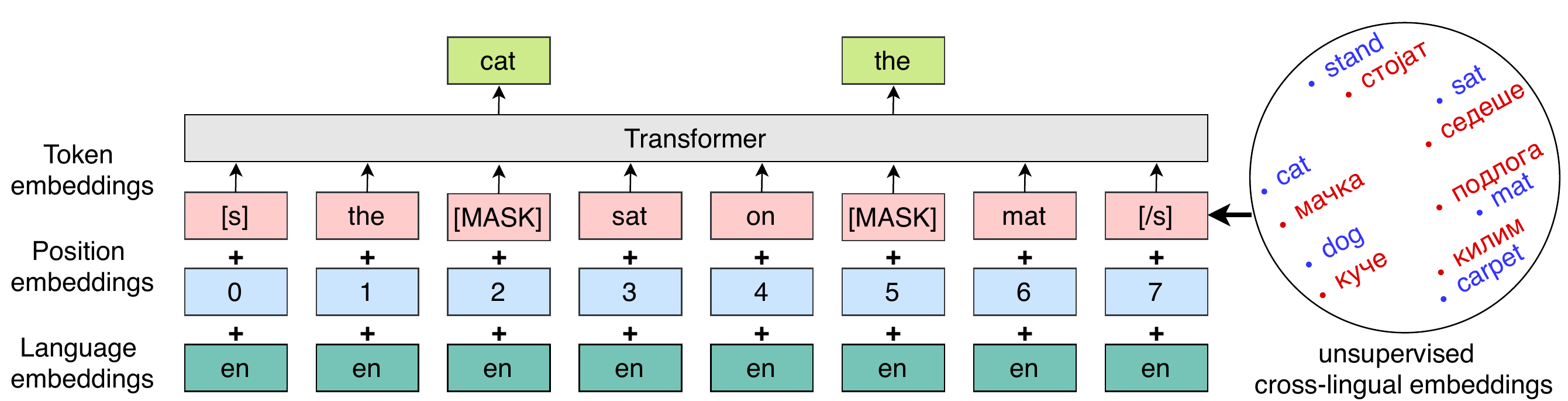}
	\caption{Lexically aligned cross-lingual masked language model.}\label{fig:lexxie}
\end{figure*}

We initialize the token embedding layer 
of the \textsc{mlm} Transformer encoder 
with pretrained \textit{VecMap} embeddings, which provide an informative mapping, i.e., cross-lingual  lexical representations. We train the model on data from both languages, using masked language modeling. Training a masked language model  enhances the cross-lingual signal by encoding  contextual representations. This step is illustrated in Figure \ref{fig:lexxie}.

\subsection{Unsupervised \textsc{nmt}} 
Finally, we transfer the \textsc{mlm}-trained encoder Transformer to an encoder-decoder translation model. We note that  the encoder-decoder attention of the Transformer is randomly initialized. We then train the model for \textsc{nmt} in an unsupervised way, using denoising auto-encoding \cite{vincent2008extracting} and back-translation \cite{sennrich2015improving}, which is performed in an online manner. This follows work by \newcite{artetxe2018unsupervised, lample2017unsupervised, lample2018phrase}.

\section{Experiments}
\label{sec:exps}
\textbf{Datasets.} We conduct experiments on English-Macedonian  (En-Mk) and English-Albanian (En-Sq), as Mk, Sq are low-resource languages, where lexical-level alignment can be most beneficial. We use 3K randomly sampled sentences of SETIMES \cite{tiedemann-2012-parallel} as validation/test sets. 
 We also use 68M En sentences from NewsCrawl. For Sq  and Mk we use all the CommonCrawl corpora from \newcite{ortizsuarez:hal-02148693}, which are $4$M Sq and $2.4$M Mk sentences.

\noindent 
\textbf{Baseline.} We use a method that relies on cross-lingual language model pretraining, namely \textsc{xlm} \cite{lample2019cross}.  This approach trains a bilingual \textsc{mlm} separately for En-Mk and En-Sq, which is used to initialize the encoder-decoder of the corresponding \textsc{nmt} system. Each system is then trained in an unsupervised way.

\noindent 
\textbf{Comparison to state-of-the-art.} 
We apply our proposed approach to \textsc{re-lm} \cite{chronopoulou2020reusing}, a state-of-the-art approach for low-resource \textsc{unmt}. This method trains a monolingual En \textsc{mlm} model (\textit{monolingual pretraining step}). Upon convergence, a vocabulary extension method is used, that randomly initializes the newly added vocabulary items. Then, the \textsc{mlm} is fine-tuned to the two languages (\textit{MLM fine-tuning step}) and used to initialize an encoder-decoder model. This method outperforms \textsc{xlm} on low-resource scenarios.  

\noindent \textbf{Lexically aligned language models.} 
When applied to the baseline, our method initializes the embedding layer of \textsc{xlm} with unsupervised cross-lingual embeddings. Then, we train \textsc{xlm} on the two languages of interest with a masked language modeling objective. Upon convergence, we transfer it to the encoder and decoder of an \textsc{nmt} model, which is trained in an unsupervised way. 

In the case of \textsc{re-lm}, our method is applied to the \textit{MLM fine-tuning step}. Instead of randomly initializing the new embedding vectors added in this step, we use pretrained unsupervised cross-lingual embeddings. We obtain them by applying \textit{VecMap} to \textit{fastText} pretrained Albanian/Macedonian embeddings and the English \textsc{mlm} token-level embeddings. Then, the \textsc{mlm} is fine-tuned on both languages. Finally, it is used to initialize an encoder-decoder \textsc{nmt} model. 


\noindent \textbf{Unsupervised \textit{VecMap} bilingual embeddings.} 
We build monolingual embeddings with the \textit{fastText} skip-gram model 
with 1024 dimensions, using our BPE-split \cite{sennrich2015neural} monolingual corpora. We map them to a shared space, using \textit{VecMap} with identical tokens. We concatenate the aligned embeddings of the two languages and use them to initialize the embedding layer of \textsc{xlm}, or the new vocabulary items of \textsc{re-lm}.
\begin{table*}[ht]
\centering
\small

\begin{tabular}{lrrrrrrrr}
\toprule
 &  \multicolumn{2}{c}{\textbf{Mk$\rightarrow$En}}  &  \multicolumn{2}{c}{\textbf{En$\rightarrow$Mk}}  
  &  \multicolumn{2}{c}{\textbf{Sq$\rightarrow$En}} &
 \multicolumn{2}{c}{\textbf{En$\rightarrow$Sq}}     \\

& BLEU $\uparrow$& \textsc{chr}F1 $\uparrow$ & BLEU $\uparrow$& \textsc{chr}F1 $\uparrow$ & BLEU $\uparrow$& \textsc{chr}F1 $\uparrow$ &BLEU $\uparrow$& \textsc{chr}F1 $\uparrow$  \\
 \midrule
 \midrule 
\textsc{\textsc{xlm}}  & 20.7 & 48.5 &  19.8 & 42.4 & 31.1 & 56.8  & 31.3  & 56.2 \\
\textit{lexically aligned  {\small{XLM}}} & 25.2 & 49.9 &  22.9 & 43.1 & 32.8 & 58.2 & 33.5 & 56.8 \\ \midrule
\textsc{re-lm}  & 25.0 & 51.1  & 23.9  & 45.8 & 30.1 & 55.8 & 32.2   & 56.4 \\
\textit{lexically aligned {\small{RE-LM}}}  & 25.3 & 51.5  &  25.6 & 47.6 & 30.5 & 56.0 & 32.9 & 56.7 \\

\bottomrule

\end{tabular}
\caption{\textsc{\textsc{unmt}} results for translations to and from English. 
The first column indicates the pretraining method used. The scores presented are significantly different
(p < 0.05) from the respective baseline. \textsc{chr}F1 refers to character n-gram F1 score \cite{popovic-2015-chrf}. The models in italics are ours.}
\label{table:results_main}
\end{table*}

\noindent \textbf{Preprocessing.} 
We tokenize the monolingual data and validation/test sets using Moses \cite{koehn2006open}.
For \textsc{xlm} \cite{lample2019cross}, we use BPE splitting with 32K operations jointly learned on both languages. For \textsc{re-lm} \cite{chronopoulou2020reusing}, we learn 32K BPEs on En for pretraining, and then 32K BPEs on both languages for the fine-tuning and \textsc{unmt} steps. The BPE merges are learned on a subset of the En corpus and the full Sq or Mk corpus. 

\noindent \textbf{Model hyperparameters.} 
We use a 
Transformer architecture for both the baselines and  \textsc{unmt} models, using the same hyperparameters as \textsc{xlm}.  For the encoder Transformer used for masked language modeling, the embedding and model size is 1024 and the number of attention heads is 8. The encoder Transformer has 6 layers, while the \textsc{nmt} model is a  6-layer encoder/decoder Transformer. The learning rate is set to $10^{-4}$ for \textsc{xlm} and \textsc{unmt}. We train the models on 8 NVIDIA GTX 11 GB GPUs. To be comparable with \textsc{re-lm}, we retrain it on 8 GPUs, as that work reports  \textsc{unmt} results with only 1 GPU. The per-GPU batch size is 32 during \textsc{xlm} and 26 during \textsc{unmt}. Our models are built on the publicly available \textsc{xlm} and \textsc{re-lm} codebases. 
We generate final translations with beam search of size 5 and
we evaluate with
SacreBLEU\footnote{Signature ``BLEU+c.mixed+\#.1+s.exp+tok.13a+v.1.4.9''}
\cite{post-2018-call}.

\vspace{-2mm} \section{Results} Table \ref{table:results_main} shows the results of our approach compared to two pretraining approaches that rely on \textsc{mlm} training, namely \textsc{xlm} and \textsc{re-lm}. The lexically aligned \textsc{xlm} improves translation results over the baseline \textsc{xlm} model. We obtain substantial improvements on En-Sq in both directions, of at most $2.2$ BLEU and $1.4$ \textsc{chr}F1, while on En-Mk, we get an even larger performance boost of up to $4.5$ points in terms of BLEU and $1.4$ in terms of \textsc{chr}F1. 
Our lexically aligned \textsc{re-lm} also consistently outperforms \textsc{re-lm}, most notably in the En$\rightarrow$Mk direction,
by up to $1.7$ BLEU. At the same time, \textsc{chr}F1 score improves by up to $1.8$ points using the lexically aligned pretraining approach compared to \textsc{re-lm}.

 In the case of \textsc{xlm}, the effect of cross-lingual lexical alignment is more evident for En-Mk, as Mk is less similar to En, compared to Sq. This is mainly the case because the two languages use a different alphabet (Latin for En and Cyrillic for Mk). 
This is also true for \textsc{re-lm} when translating out of En, showing that
enhancing the \textit{fine-tuning step} of \textsc{mlm} with pretrained embeddings is  helpful and improves the final \textsc{unmt} performance.
 
In general, our method provides better alignment of the lexical-level representations of the \textsc{mlm}, thanks to the transferred \textit{VecMap} embeddings.
We hypothesize that static cross-lingual embeddings enhance the  knowledge that a cross-lingual masked language model obtains during training. As a result, using them to bootstrap the pretraining procedure improves the ability of the model to map the distributions of the two languages and yields higher translation scores. Overall, our approach consistently outperforms two pretraining models for \textsc{unmt}, providing for the highest BLEU and \textsc{chr}F1 scores on all translation directions.

\begin{table}[h]
\centering
\small

\begin{tabular}{lrrrr}
\toprule
 &    \multicolumn{2}{c}{\textbf{En-Mk}}          &  \multicolumn{2}{c}{\textbf{En-Sq}}     \\
 &   NN & CSLS          &   NN & CSLS    \\
\midrule
\midrule

\textsc{\textsc{xlm}} & 6.3 & 6.5 & 43.0 & 40.7 \\
lexically aligned  \textsc{\textsc{xlm}} & \textbf{15.5} & \textbf{16.5} &\textbf{51.6} & \textbf{50.6} \\ 
\midrule
\textsc{re-lm} & 29.8 & 16.1 & 52.0  & 35.9 \\
lexically aligned \textsc{re-lm} & \textbf{32.0} & \textbf{17.2} & \textbf{53.0} & \textbf{36.9} \\ 

\bottomrule

\end{tabular}
\caption{P@5 results for the BLI task on the MUSE \cite{conneau2017word} dictionaries. We evaluate the alignment of the embedding layer of each trained \textsc{mlm}. 
}
\label{table:cosinesim}
\end{table}


\section{Analysis}

We conduct an analysis to assess the contribution of lexical-level alignment in the \textsc{mlm} training. We present Bilingual Lexicon Induction (BLI) and BLEU 1-gram precision scores. We also investigate the best method to leverage pretrained cross-lingual embeddings during \textsc{mlm} training, in terms of final \textsc{unmt} performance. 


\noindent \textbf{Bilingual Lexicon Induction (BLI).} 
We use BLI, a standard way of evaluating lexical quality of embedding representations \cite{pmlr-v37-gouws15,survey-crosslingual}, to explore the effect of the alignment of our method.
We compare the BLI score of different cross-lingual pretrained language models.  
We report precision@5 (P@5) using nearest neighbors (NN) and cross-lingual semantic similarity (CSLS).
The results are presented in Table \ref{table:cosinesim}. We use the embedding 
layer of each \textsc{mlm} for this task. We also experimented with averages over different layers, but noticed the same trend in terms of BLI scores.
We obtain word-level representations by averaging over the corresponding subword embeddings. It is worth noting that we compute the type-level representation of each vocabulary word in isolation, similar to \citet{Vulic2020ProbingPL}.

In Table \ref{table:cosinesim}, we observe that lexical alignment is more beneficial for En-Mk. This can be explained by the limited vocabulary overlap of the two languages, which does not provide sufficient  cross-lingual signal for the training of \textsc{mlm}. 
By contrast, initializing an \textsc{mlm} with pretrained embeddings largely improves performance, even for a higher-performing model, such as \textsc{re-lm}. In En-Sq, the effect of our approach is smaller yet consistent. This can be attributed to the fact that the two languages use the same script.  

Overall, our method enhances the lexical-level information captured by pretrained \textsc{mlm}s, as shown empirically. This is consistent with our intuition that cross-lingual embeddings capture a bilingual signal that can benefit \textsc{mlm} representations.

\begin{table}
\centering
\small

\begin{tabular}{lrrrr}
\toprule
 &    \multicolumn{2}{c}{\textbf{En-Mk}}    & \multicolumn{2}{c}{\textbf{En-Sq}}     \\
 &   $\leftarrow$ & $\rightarrow$         &   $\leftarrow$ & $\rightarrow$          \\
\midrule
\midrule
\textsc{\textsc{xlm}} & 53.1 & 41.4 &  62.1 & 60.4  \\
lexically aligned  \textsc{\textsc{xlm}} & \textbf{56.0} & \textbf{51.8} & \textbf{63.6} & \textbf{61.5} \\ 
\midrule
\textsc{re-lm} & 56.0 & 52.8 & 61.6 & 61.2 \\
lexically aligned \textsc{re-lm} & \textbf{56.6}  & \textbf{53.9} & \textbf{62.0} & \textbf{61.7} \\ 

\bottomrule

\end{tabular}
\caption{BLEU 1-gram precision scores. 
} 

\label{table:unigram}
\end{table}
\noindent \textbf{1-gram precision scores.}
To examine whether the improved translation performance is a result of the  lexical-level information provided by static embeddings, we present 1-gram precision scores in Table \ref{table:unigram}, as they can be directly attributed to lexical alignment. The biggest performance gains (up to +$10.4$) are obtained when the proposed approach is applied to \textsc{xlm}. This correlates with the BLEU scores of Table \ref{table:results_main}. Moreover, the En-Mk language pair benefits more than En-Sq from the lexical-level alignment both in terms of 1-gram precision and BLEU. These results show that the improved BLEU scores can be attributed to the enhanced lexical representations. 

\begin{table}[h]
\centering

\resizebox{\columnwidth}{!}{
\begin{tabular}{lrrrr}
\toprule
Alignment Method &  \multicolumn{2}{c}{\textbf{En-Mk}}        & \multicolumn{2}{c}{\textbf{En-Sq}}     \\
  &  $\leftarrow$  &  $\rightarrow$      &   $\leftarrow$  &  $\rightarrow$     \\
    
 \midrule
 \midrule 
lexically aligned \textsc{mlm} & & & & \\  
\hspace{4mm} frozen embeddings   & 24.7 & 22.1 & 31.0 & 32.1  \\
\hspace{4mm} fine-tuned embeddings (ours)  & \textbf{25.2} &  \textbf{22.9} &  \textbf{32.8} &  \textbf{33.5} \\

\bottomrule

\end{tabular}}
\caption{BLEU scores using different initializations of the \textsc{xlm} embedding layer. \textsc{xlm} is then trained on the respective language pair and used to initialize a \textsc{unmt} system. Both embeddings are aligned using \textit{VecMap}. 
}
\label{table:freeze}
\end{table}
\noindent \textbf{How should static embeddings be integrated in the \textsc{mlm} training?} 
We explore different ways of incorporating the lexical knowledge of pretrained cross-lingual embeddings to the second, masked language modeling stage of our approach (\S\ref{sec:mlm}). Specifically, we keep the aligned embeddings fixed (\textit{frozen}) during \textsc{xlm} training and compare the performance of the final \textsc{unmt} model to the proposed (\textit{fine-tuned}) method. We point out that, after we transfer the trained \textsc{mlm} to an encoder-decoder model, all layers are trained for \textsc{unmt}.

Table \ref{table:freeze} summarizes our results. 
The fine-tuning approach, which is adopted in our proposed method, provides a higher performance both in En-Mk and En-Sq, with the improvement being more evident in En-Sq. Our findings generally show that it is preferable to train the bilingual embeddings together with the rest of the model in the \textsc{mlm} step.


\section{Related Work}
\citet{artetxe2017unsupervised,lample2017unsupervised} initialize \textsc{unmt} models with word-by-word translations, based on a bilingual lexicon inducted in an unsupervised way by the same monolingual data, or simply with cross-lingual embeddings.
\citet{lample2018phrase} also use pretrained embeddings, learned on joint monolingual corpora of the two languages of interest, to initialize the embedding layer of the encoder-decoder.
 \citet{lample2019cross} remove pretrained embeddings from the \textsc{unmt} pipeline and align language distributions by simply pretraining a \textsc{mlm} on both languages, in order to learn a cross-lingual mapping. 
However, it has been shown that this pretraining method provides a weak alignment of the language distributions \cite{ren-etal-2019-explicit}. While that work identified as a cause the lack of sharing n-gram level cross-lingual information, we address the lack of cross-lingual information at the lexical level.  

Moreover, most prior work on \textsc{unmt} focuses on languages with abundant, high-quality monolingual corpora. In low-resource scenarios though, especially when the languages are not related, pretraining a cross-lingual \textsc{mlm} for unsupervised \textsc{nmt} does not yield good results \cite{guzman2019flores, chronopoulou2020reusing}. We propose a method that overcomes this issue by enhancing the \textsc{mlm} with cross-lingual lexical-level representations.

Another line of work tries to enrich the representations of multilingual \textsc{mlm}s with additional knowledge \cite{wang2020kadapter, pfeiffer-etal-2020-mad}  without harming the already-learned representations. In our work, we identify lexical information as a source of knowledge that is missing from \textsc{mlm}s, especially when it comes to low-resource languages. Surprisingly, static embeddings, such as \textit{fastText}, largely outperform representations extracted by multilingual \textsc{mlm}s in terms of cross-lingual lexical alignment \cite{Vulic2020ProbingPL}. Motivated by this, we aim to narrow the gap between the lexical representations of bilingual \textsc{mlm}s and static embeddings, in order to achieve a higher translation quality, when transferring the \textsc{mlm} to an encoder-decoder \textsc{unmt} model. 

\section{Conclusion}
We propose a method to improve the lexical ability of a Transformer encoder by initializing its embedding layer with pretrained cross-lingual embeddings. The Transformer is trained for masked language modeling on the language pair of interest. After that, it is used to initialize an encoder/decoder model, which is trained for \textsc{unmt} and outperforms relevant baselines. Results confirm our intuition that masked language modeling, which provides contextual representations,  benefits from cross-lingual embeddings, which capture lexical-level information.  In the future, we would like to investigate whether lexical knowledge can be infused to  multilingual \textsc{mlm}s. We would also like to experiment with other schemes of training the \textsc{mlm} in terms of how the embedding layer is updated, such as regularizer annealing strategies, which would enable keeping the embeddings relatively fixed, but still allow for some limited training.

\section{Ethical Considerations}

In this work, we propose a novel unsupervised neural machine translation approach, which is tailored to low-resource languages in terms of monolingual data. We experiment with unsupervised translation between English, Albanian and Macedonian. 

For English, we use high-quality data from news articles. The Albanian and Macedonian monolingual data originates from the OSCAR project \cite{ortizsuarez:hal-02148693}. The corpora are shuffled and stripped of all metadata. Therefore, the data should not be easily attributable to specific individuals. Nevertheless, the project offers easy ways to remove data upon request. The En-Sq and En-Mk parallel development and test data are obtained from OPUS \cite{tiedemann-2012-parallel} and consist of high-quality news articles.

Our work is partly based on training type-level embeddings which are not computationally expensive. 
However, training cross-lingual masked language models requires significant computational resources. To lower environmental impact, we do not conduct  hyper-parameter search and use well-established values for all hyper-parameters.

\section*{Acknowledgments}

This project has received funding from the European Research Council under the European Union’s Horizon $2020$ research and innovation program  (grant agreement 
\#$640550$). This work was also supported by DFG (grant FR $2829$/$4$-$1$). We thank Katerina Margatina,  Giorgos Vernikos and Viktor Hangya for their thoughtful comments/suggestions and valuable feedback. 


\bibliography{anthology,custom}
\bibliographystyle{acl_natbib}

\clearpage
\appendix

\section{Appendix}
\label{sec:appendix}

\subsection{Datasets} 
 We remove sentences longer than $100$ words after BPE splitting. We split the data using the fastBPE codebase\footnote{\url{https://github.com/glample/fastBPE}}. 
 
 \subsection{Model Configuration}
 We tie the embedding and output (projection) layers of both \textsc{lm} and \textsc{nmt} models \cite{press-wolf-2017-using}. We use a dropout rate of $0.1$ and  GELU activations \cite{hendrycks2016bridging}. We use the default parameters of \citet{lample2019cross} in order to train our models.
 
Regarding the runtimes for the En-Sq experiments: the baseline \textsc{xlm} was trained for 3 days on 8 GPUs while our approach for 6 days and 14h. The experiment with freezing the embeddings provided for faster training, 2 days and 17h. The three methods needed 23h, 21h, and 1d and 8h for the \textsc{unmt} part, respectively. 
Fine-tuning with \textsc{re-lm} took 2 days and 14h on 1 GPU and with our approach it took 1 day and 5h. \textsc{unmt} for these models took 2 days and 11h, and 13h, respectively. 
We get a checkpoint every $50$K sentences processed by the model.

 \subsection{Validation Scores of Results}

In Table \ref{table:results_main_dev} we show the dev scores of the main  results, in terms of BLEU scores. This table extends Table \ref{table:results_main} of the main paper. 

In Table \ref{table:freeze-dev}, we show the dev scores of the extra fine-tuning experiments we did for the analysis. The table corresponds to Table \ref{table:freeze} of the main paper. 

\begin{table}[ht]
\centering
\small

\begin{tabular}{lrrrr}
\toprule
 &  \multicolumn{2}{c}{\textbf{En-Mk}}        & \multicolumn{2}{c}{\textbf{En-Sq}}     \\
  &  $\leftarrow$  &  $\rightarrow$      &   $\leftarrow$  &  $\rightarrow$     \\
    
 \midrule
 \midrule 
\textsc{\textsc{xlm}}    & - & -  & 30.7 & 32.0  \\
lexically aligned  \textsc{\textsc{xlm}}    &\textbf{24.6} &  \textbf{23.3} &  \textbf{31.9} &  \textbf{33.8} \\ \midrule
\textsc{re-lm}  & 25.0  & 25.7  & 29.9 & 32.8 \\
lexically aligned \textsc{re-lm}   & \textbf{25.3} & \textbf{26.6}  & \textbf{29.5}  & \textbf{30.3}  \\

\bottomrule

\end{tabular}
\caption{\textsc{\textsc{unmt}} BLEU scores on the development set. 
}
\label{table:results_main_dev}
\end{table}

\begin{table}[ht!]
\centering

\resizebox{\columnwidth}{!}{
\begin{tabular}{lrrrr}
\toprule
Alignment Method &  \multicolumn{2}{c}{\textbf{En-Mk}}        & \multicolumn{2}{c}{\textbf{En-Sq}}     \\
  &  $\leftarrow$  &  $\rightarrow$      &   $\leftarrow$  &  $\rightarrow$     \\
    
 \midrule
 \midrule 
lexically aligned \textsc{mlm} & & & & \\  
\hspace{4mm} frozen embeddings   & \textbf{24.8} & 23.0 & 31.0 & 32.1  \\
\hspace{4mm} fine-tuned embeddings (ours)  &24.6 &  \textbf{23.3} &  \textbf{31.1} &  \textbf{32.0} \\

\bottomrule

\end{tabular}}
\caption{Development BLEU scores using different initializations of the \textsc{xlm} embedding layer.
}
\label{table:freeze-dev}
\end{table}
We note that the dev scores are obtained using greedy decoding, while the test scores are obtained with beam search of size $5$. We clarify that we train each \textsc{nmt} model using as training criterion the validation BLEU score of the Sq, Mk$\rightarrow$En direction, with a patience of $10$. We specifically use the \texttt{multi-bleu.perl} script from Moses. 

\subsection{\textit{Joint} \textit{vs} \textit{VecMap} embeddings.} 

Using joint embeddings to initialize the \textsc{mlm}, before training it on data from the respective language is less effective for \textsc{unmt}. This is mostly the case for En-Mk, since the two languages use a different alphabet (Latin and Cyrillic). In this case, simply learning \textit{fastText} embeddings on the concatenation of the two corpora is not useful, because the languages do not have a big lexical overlap. 

\begin{table}[ht!]
\centering
\small

\resizebox{\columnwidth}{!}{
\begin{tabular}{lrrrr}
\toprule
Alignment Method &  \multicolumn{2}{c}{\textbf{En-Mk}}        & \multicolumn{2}{c}{\textbf{En-Sq}}     \\
  &  $\leftarrow$  &  $\rightarrow$      &   $\leftarrow$  &  $\rightarrow$     \\
    
 \midrule
 \midrule 
joint \textit{fastText}  & 21.5 & 19.8 &  32.3  & 33.1 \\

\textsc{VecMap}    &\textbf{25.2} &  \textbf{22.9} &  \textbf{32.8} &  \textbf{33.5}\\

\bottomrule

\end{tabular}}
\caption{BLEU scores using different initializations of the \textsc{xlm} embedding layer. 
\textsc{xlm} is then trained on the respective language pair and used to initialize a \textsc{unmt} system. 
\textit{Joint fastText embs} refers to jointly learned  embeddings following \citet{lample2018phrase}.
}
\label{table:freeze2}
\end{table}

\end{document}